\documentclass[10pt,twocolumn,letterpaper]{article}

\usepackage{cvpr}              

\usepackage{graphicx}
\usepackage{amsmath}
\usepackage{amssymb}
\usepackage{booktabs}

\usepackage{multicol}
\usepackage{multirow}
\usepackage{color}
\usepackage{makecell}
\usepackage{booktabs}
\usepackage{tablefootnote}
\usepackage{color, colortbl}
\definecolor{Gray}{gray}{0.93}
\definecolor{gray}{gray}{0.5}

\usepackage[pagebackref,breaklinks,colorlinks]{hyperref}

\usepackage[capitalize]{cleveref}
\crefname{section}{Sec.}{Secs.}
\Crefname{section}{Section}{Sections}
\Crefname{table}{Table}{Tables}
\crefname{table}{Tab.}{Tabs.}

\begin{document}

\title{Learning Grounded Vision-Language Representation for Versatile Understanding in Untrimmed Videos}

\author{Teng Wang$^{12*}$, Jinrui Zhang$^{1*}$, Feng Zheng$^{1\dagger}$, Wenhao Jiang$^3$, Ran Cheng$^1$, Ping Luo$^2$\\
$^1$Southern University of Science and Technology, $^2$The University of Hong Kong, $^3$Tencent
}
\maketitle

\begin{abstract}
\let\thefootnote\relax\footnotetext{$^*$ Equal contribution $^\dagger$ Corresponding author}

Joint video-language learning has received increasing attention in recent years. However, existing works mainly focus on single or multiple trimmed video clips (events), which makes human-annotated event boundaries necessary during inference. To break away from the ties, we propose a grounded vision-language learning framework for untrimmed videos, which automatically detects informative events and effectively excavates the alignments between multi-sentence descriptions and corresponding event segments. Instead of coarse-level video-language alignments, we present two dual pretext tasks to encourage fine-grained segment-level alignments, i.e., text-to-event grounding (TEG) and event-to-text generation (ETG). TEG learns to adaptively ground the possible event proposals given a set of sentences by estimating the cross-modal distance in a joint semantic space. Meanwhile, ETG aims to reconstruct (generate) the matched texts given event proposals, encouraging the event representation to retain meaningful semantic information. To encourage accurate label assignment between the event set and the text set, we propose a novel semantic-aware cost to mitigate the sub-optimal matching results caused by ambiguous boundary annotations. Our framework is easily extensible to tasks covering visually-grounded language understanding and generation. We achieve state-of-the-art dense video captioning performance on ActivityNet Captions, YouCook2 and YouMakeup, and competitive performance on several other language generation and understanding tasks. Our method also achieved 1st place in both the MTVG and MDVC tasks of the PIC 4th Challenge. Our code is publicly available at \url{https://github.com/zjr2000/GVL}.
\end{abstract}

\section{Introduction}

\begin{figure}
  \includegraphics[width=0.5\textwidth]{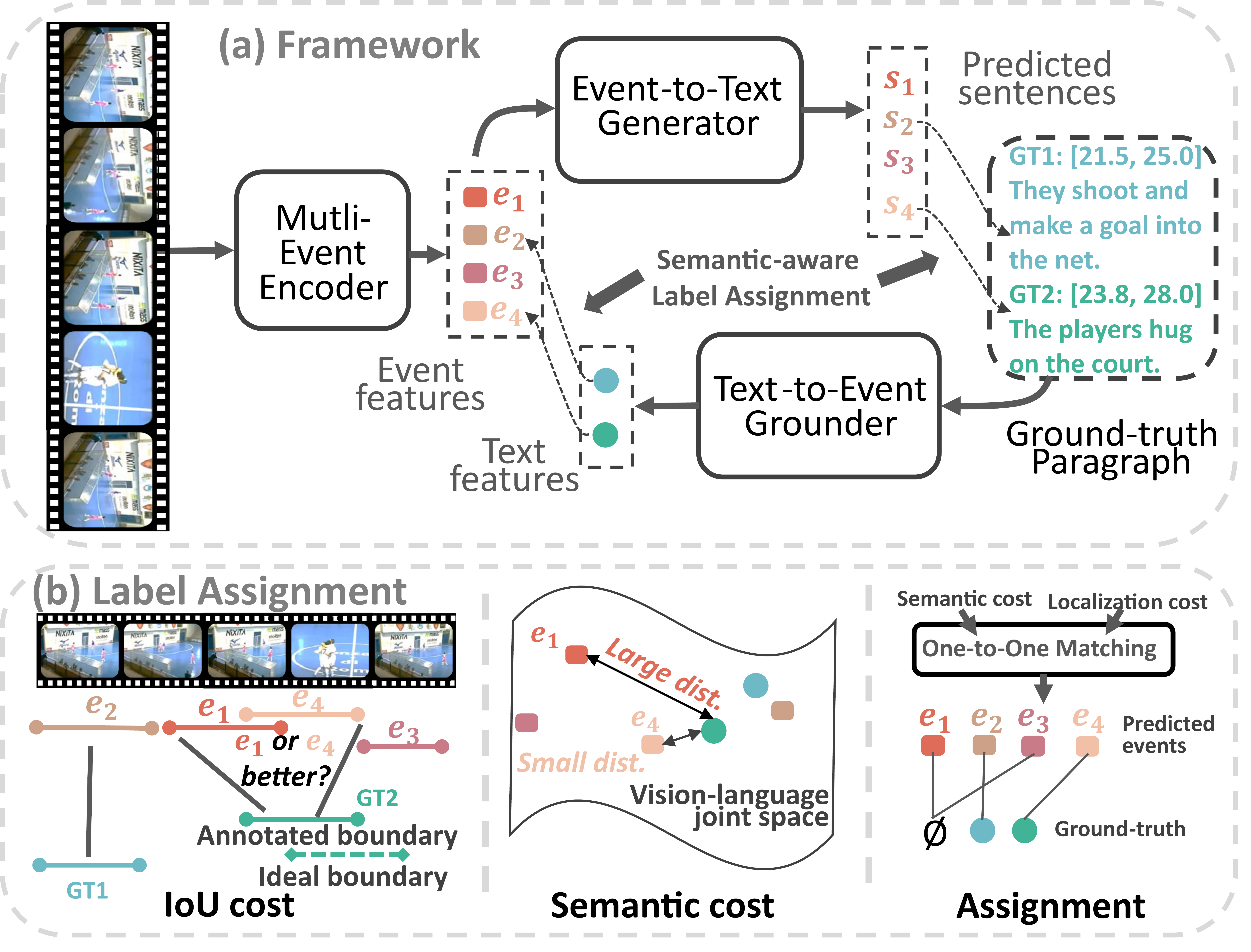}
  \caption{Illustration of the proposed grounded vision-language learning framework. (a) Two dual pretext tasks are proposed to exploit the fine-grained correspondence between the event set and the sentence set, encouraging the event encoder to learn semantically rich and temporally-sensitive features. (b) For label assignment between detected events and ground-truth sentences, IoU cost (or other localization-aware costs) causes sub-optimal matching results when segment annotations are ambiguous, specifically for candidate events with similar IoU. The proposed semantic-aware cost could address the problem by considering the event-sentence similarity in a joint semantic space as the matching criterion.}
  \vspace{-1em}
  \label{fig:intro}
\end{figure}

Over the past few years, advances in deep learning have led to signiﬁcant progress in video understanding~\cite{karpathy2014large, tran2015learning, wang2018temporal, arnab2021vivit, geng2022spatial}. Typically, video recognition models are optimized for high classification accuracy on a discrete set of action categories. Such a paradigm could achieve surprising performance on predefined categories but impede the transferability to recognize the novel semantic concepts. Recently, CLIP~\cite{radford2021learning} and its variants~\cite{radford2021learning, luo2021clip4clip} showed that joint vision-language (VL) modeling could learn visual representations that are semantic-rich and transferable to diverse downstream tasks, across vision tasks~\cite{patashnik2021styleclip, li2021grounded} and VL tasks~\cite{shen2021much}.

There has been a rich study~\cite{sun2019videobert, lei2021less, xu2021videoclip} focusing on joint learning between a video and a sentence. However, video datasets used in those methods are well-trimmed by human annotators, ensuring that each video contains only a single event. To model long videos, some works~\cite{zhang2018cross, ging2020coot} tackle cross-modal interactions between multiple event segments and a paragraph. However, their inputs are multiple trimmed segments, which means they also require event boundary annotations during inference, significantly limiting their applications in real-world scenarios.

This paper focuses on joint VL modeling for long, untrimmed videos. Given an untrimmed video containing background frames and a paired paragraph (\ie, multiple sentences with their temporal locations\footnote{The inference stage of our method is free of location annotations while the model still needs them for cross-modal correspondence during training.}), we aim to learn event-level, temporally-sensitive, and semantically-rich visual representations from visually-grounded language. We identify two conceptual challenges: (1) As the semantic information is not spread evenly across time in untrimmed videos, learning event-level representation that could distinguish the background segments from foreground ones is difficult. The mainstream paradigm for localization tasks learns event features from action labels. However, coarse-level supervision signals lack discriminative of different events occurring in the same video since these events typically fall into the same action class. (2) Current untrimmed video-language datasets (\eg, ActivityNet Captions~\cite{krishna2017dense}) contain ambiguous annotations of event boundaries, which is mainly caused by the subjective bias of annotators~\cite{otani2020uncovering,huang2022video,chen2020refinement}. It is less explored how to assign ground-truth annotations with noisy boundary annotations to predicted events. The privileged solution for label assignment in visual grounding~\cite{wang2021negative, kamath2021mdetr} utilizes localization similarity (\eg, Intersection over Union) as the matching cost. As shown in Fig.~\ref{fig:intro} (bottom), localization cost may produce sub-optimal matching results for sentences with inaccurate boundary annotations, especially when several predicted events have a similar IoU with the same ground-truth sentence.

To address the above challenges, we propose to learn uni-modal event-level representation from language supervision by solving a bidirectional set prediction problem. 
Specifically, two dual pretext tasks are presented as the supervision for an event localizer, \ie, text-to-event grounding (TEG) and event-to-text generation (ETG). TEG learns to ground the corresponding events given sentences, which 
requires the event representation to be discriminative enough so that text queries can distinguish the matched events from negative ones. Meanwhile, ETG learns to transform the event set into a sentence set, ensuring that event features preserve rich semantics as much as possible to regenerate a natural language description. 

Furthermore, considering that both TEG and ETG objectives rely on the correspondence between predicted events and ground truth, we propose a robust label assignment strategy based on semantic similarity for set-level matching. Different from localization costs that are low-level cues and easily affected by noisy annotations, the semantic similarity between predicted events and ground truth serves as a more robust judgment, which could alleviate the negative effect of ambiguous annotations and makes friendly matching decisions toward fine-grained video understanding.

We conduct extensive experiments on four VL tasks on four untrimmed video benchmarks, \ie, ActivityNet Captions~\cite{krishna2017dense}, TACoS~\cite{regneri2013grounding}, YouMakeup~\cite{wang2019youmakeup} and YouCook2~\cite{zhou2018towards}. Quantitive and qualitative experiments both verify that the model learns event-level, temporally discriminative, and semantic-rich visual features for untrimmed videos. 

The contributions of this paper are three-fold: 1) A new vision-language framework for untrimmed videos is proposed by solving a bidirectional set prediction problem, in which fine-grained cross-modal alignment between the predicted events and visually-grounded sentences is exploited. 
2) A novel semantic-aware matching cost is proposed to mitigate the noise in boundary annotations, enabling a more reliable label assignment between events and ground truth than traditional localization cost. 
3) Extensive experiments on four video benchmarks show that our model could achieve state-of-the-art performance on dense video captioning and competitive performance on three other vision-language understanding and generation tasks.

\section{Related Work}

\begin{figure*}
\centering
  \includegraphics[width=0.95\textwidth]{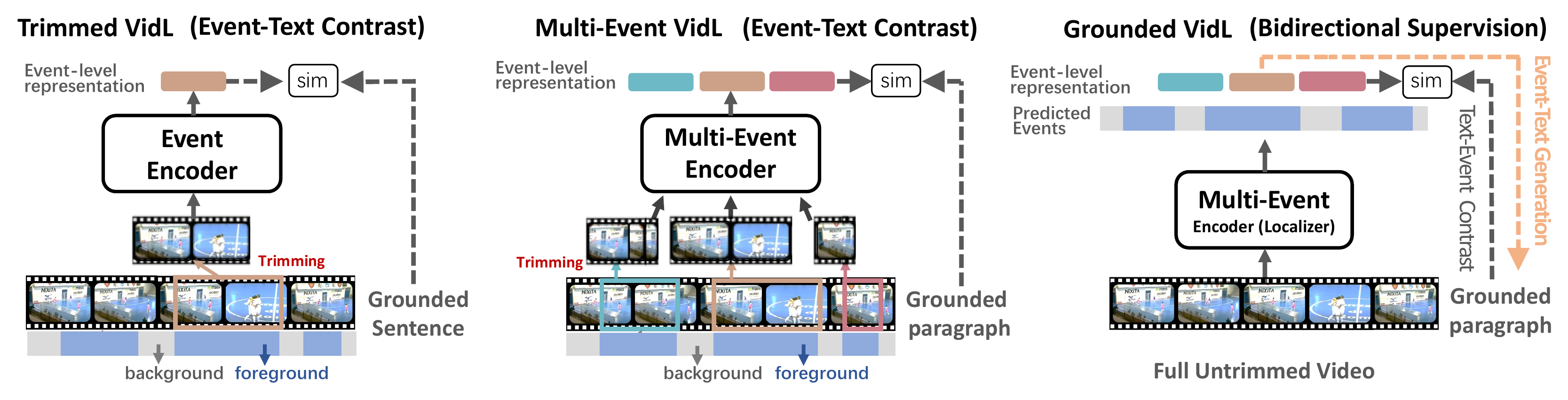}
    \caption{Illustration of dual-stream video-language (VidL) representation learning frameworks. Trimmed (\eg,~\cite{bain2021frozen,ge2022bridging}) and multi-event VidL learning (\eg,~\cite{ging2020coot,xu2021videoclip}) take the single trimmed clip (event) or multiple trimmed clips as inputs. They lack the perception of long-term video dynamics across continuous video streams, limiting their application in temporal localization tasks.
    The proposed grounded VidL framework takes the full untrimmed video as input (both foreground and background frames), learning to localize salient event-level features from the supervision of bidirectional cross-modal interactions. 
The learned features are encouraged to be temporal-sensitive for distinguishing foreground/background frames, meanwhile semantically rich for fine-grained VidL understanding or generation tasks.}
  \vspace{-1em}
  \label{fig:methods_diff}
\end{figure*}

\subsection{Video-Language Representation Learning}
Recently, a variety of approaches for video language (VidL) representation learning has been proposed. These methods can be broadly divided into two categories, depending on whether an early fusion strategy is adopted or not. Single-stream methods~\cite{zhu2020actbert, sun2019videobert, xu-etal-2021-vlm} typically input video and text directly into a single cross-modal encoder to model their interactions. Dual-stream methods~\cite{ging2020coot, bain2021frozen, ge2022bridging, zhang2018cross, lei2021less, xu2021videoclip, li-etal-2020-hero,rouditchenko2020avlnet,miech2020end} model video and text separately and connect them through a contrastive alignment loss or cross-modal encoder. Compared with the former, the independent modeling of two modalities is more suitable for tasks with the single-modal input. 

Dominant dual-stream approaches usually truncate the untrimmed video into single~\cite{bain2021frozen,ge2022bridging,rouditchenko2020avlnet,miech2020end} or multiple~\cite{ging2020coot,xu2021videoclip,zhang2018cross} short clips with human-annotated event boundaries~\cite{caba2015activitynet} or ASR transcripts~\cite{miech2019howto100m}. The incomplete, disjointed video clips impede the learning of long-term temporal dependency between sequential events and the discriminability between foreground/background segments. 
Therefore, they prefer downstream tasks that human-annotated event boundaries are given, \eg, trimmed video recognition~\cite{bain2021frozen}, clip-level video-text retrieval~\cite{bain2021frozen,xu2021videoclip}, and video paragraph captioning~\cite{ging2020coot}, but is not applicable (or underperform) on temporal localization tasks requiring long-term reasoning, like video grounding and dense captioning.

We focus on temporal localization tasks in untrimmed videos by learning a multi-event localizer that distinguishes foreground events from background frames. The event-level representation is supervised by language from bidirectional information flow to enhance the representing ability. We illustrate the difference between representative dual-stream video-language learning paradigms in Fig.~\ref{fig:methods_diff}.

\subsection{Temporal Localization Tasks} \label{sec:task}

Our framework focuses on grounded event-level representation learning, which could be applied to several temporal localization tasks: temporal action localization (TAL), dense video captioning (DVC) and single/multi-sentence video grounding (SSVG/MSVG). TAL and DVC aim to localize all events in an untrimmed video, followed by describing the events by class labels and sentences, respectively. Two-stage methods~\cite{chao2018rethinking, heilbron2016fast, yuan2016temporal, zhao2017temporal, wang2018bidirectional, zhou2018end, Mun2019stream} firstly parse a video into event proposals and then classify/caption the confident proposals, while one-stage methods~\cite{lin2017single, alwassel2018action, wang2021end} perform proposal generation and describing simultaneously. 

SSVG/MSVG aims to localize events given single/mul-tiple text queries. Regression-based methods~\cite{yuan2019find, yuan2019semantic, chen2020rethinking, wang2020temporally} adopt an early fusion strategy for videos and texts and then regress the boundaries based on the fused features. Detection-based methods~\cite{chen2018temporally, gao2017tall, zhang2020learning, zeng2020dense} first localize all candidate events by a unimodal localizer and then perform multimodal fusion for assessing the final confidence. Metric learning-based methods~\cite{anne2017localizing, wang2021negative} project the candidate events and the texts into a vision-language joint space by two independent encoders, where separate representations are learned for each modality under metric learning constraints. 

Captioning and grounding tasks are closely related to each other in nature. While most methods tackle the dense captioning and grounding tasks separately, there are a few efforts made toward the joint learning between captioning and grounding tasks. \cite{nguyen2018weakly, chen2021towards} tackle the problem of weakly supervised dense video captioning and propose two core modules, sentence localizer and caption generator, which run in turn to extract the correspondence between event locations and target sentences. However, the learned event representations from their sentence localizers are conditioned on cross-modal inputs, limiting their usability to a broader range of vision-only localization tasks. Our method differs from previous methods:  1) We focus on learning event-level representations with only visual inputs, disentangling the event localizer from cross-modal interactions, and allowing flexible adaptation to other visual localization tasks; 2) Unlike location-oriented label assignment, we propose a semantic-aware cost that matches informative events to sentences with similar semantics for mitigating the influence of unreliable boundary annotations.

\section{Method}

\begin{figure*}
\centering
\vspace{2em}
  \includegraphics[width=0.95\textwidth]{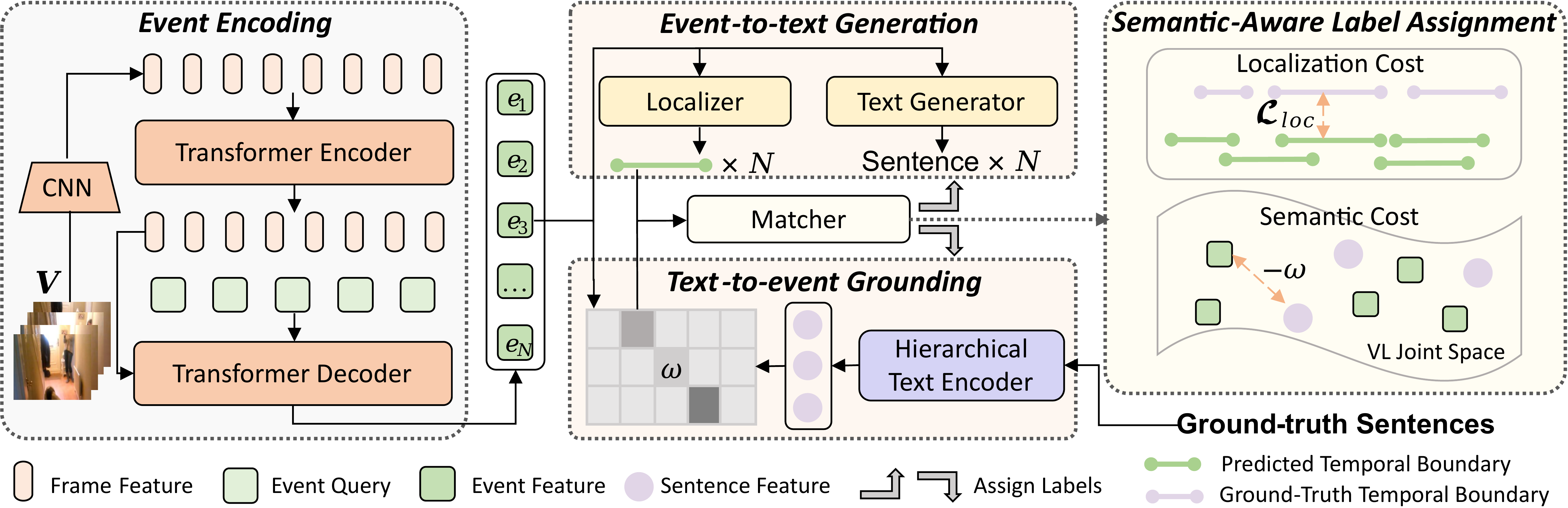}
  \caption{Illustration of the proposed grounded vision-language learning framework. The input video and paragraph are encoded by two independent encoders to get a separate feature space for two modalities.  
  The unimodal representations interact with each other by addressing two dual problems.   Text-to-event grounding aligns multiple sentences with event proposals in a one-to-one manner. Event-to-text generation decodes the event proposals into natural language. 
For better label assignment, we propose a semantic-aware matching strategy between ground-truth events and predicted proposals, which could enrich the prediction of event features with more semantics.}
  \label{fig:fig2}
\end{figure*}

This paper aims to learn event-level video representation from language descriptions by bidirectional prediction tasks. The overall framework is illustrated in Fig.~\ref{fig:fig2}. The event-level representation is encoded by a transformer-based event encoder. To ensure that learned event representation carries informative and discriminative features, we introduce bidirectional pretext tasks to exploit the cross-modal supervisions from natural language (described in Sec.~\ref{sec:bi_optimization}). To find the corresponding ground truth for each predicted event and supervise the proposal sequence as a whole, we perform a semantic-aware label assignment that is robust to noisy annotations (described in Sec.~\ref{sec:SAM}).

\subsection{Visual Learning from Grounded Language}
\label{sec:problem_formulation}

Traditional action recognition or action localization methods typically learn visual features from coarse-level action classes, which lack fine-grained discriminability to distinguish intra-class visual differences, especially under a complex scenario in long videos. We instead introduce the semantically-dense natural language as the supervision since the paragraph description for untrimmed videos carries more visual concepts and complex temporal relationships, forming a more compact label space for better representation learning.

We denote an untrimmed video as $\mathbf{V}$ and its sentence annotation as $\mathbf{Y} = \{y_k = (s_{k}, t_{k})\}_{k=1}^K$, where $s_k$ and $t_k$ are a natural sentence and its temporal location in the video. We learn an event encoder (localizer) to detect salient event features $\mathbf{E}$ under the supervision of a bidirectional information flow: Event-to-text generation (ETG) for preserving rich semantic information; Text-to-event grounding (TEG) for improving discriminability. The bidirectional learning problem is formulated as follows: 
\begin{itemize}
\itemsep -0.2em
    \item \textbf{Event Encoding}: Learning to propose a set of event-level features from an untrimmed video $f_{\theta}\!:\! \mathbf{V} \!\rightarrow \!\mathbf{E}$. 
    \item \textbf{ETG}: Learning the cross-modal mapping $g_{\theta_1}\!: \! \mathbf{E} \!
 \rightarrow \! \mathbf{Y}$.
    \item \textbf{TEG}: Learning the cross-modal mapping $h_{\theta_2}\!: \! \mathbf{Y} \! \rightarrow \! \mathbf{E}$.
\end{itemize}
Note that $\mathbf{E}=\{e_i\}_{i=1}^N$ where $e_i$ is the features of a predicted event. Considering that $g_{\theta_1}$ and $h_{\theta_2}$ both perform a set-to-set mapping, the learning process needs a label assignment strategy to assign an element from $Y$/$E$ as the learning target of $g_{\theta_1}(e_i)$ / $h_{\theta_2} ({y}_k)$ for all $e_i$/$y_k$. A reasonable label assignment  for each element in two sets is essential to ensure the  consistency of bidirectional prediction. After label assignment, the overall objective is to minimize the sum of two prediction losses:
\begin{equation}
\! \mathop{\rm minimize}\limits_{\theta, \theta_1, \theta_2}\left[\mathcal{L}_{\rm etg} \left( g_{\theta_1}(\mathbf{E}), \mathbf{Y} \right) \!+\! \alpha\mathcal{L}_{\rm teg} \left( h_{\theta_2}(\mathbf{Y}), \mathbf{E} \right) \right],
\label{eqn:loss}
\end{equation}
where $\alpha$ is a trade-off factor.

\subsection{Bidirectional Prediction Task}
\label{sec:bi_optimization}
Two complementary prediction tasks are proposed to exploit the bidirectional mapping between the event and the sentence sets in a video. The proposed TEG aims to ground corresponding events from the event set given sentence queries, focusing on discriminative features that make semantic-similar negative events more separable. It is suitable for coarse-level understanding tasks, like action detection or video grounding. However, for fine-grained generation tasks, like video captioning, discriminative features may be insufficient to generate a meaningful sentence since the generation task requires detailed visual semantics for word-level fine-grained prediction. Therefore, TEG is further introduced to enrich the model's capability to encode detailed semantics. Unlike previous works focusing on task-specific features in untrimmed videos~\cite{wang2021end, wang2021negative, zhang2020learning, zeng2020dense}, we leverage set-level relationships between visual events and textual sentences from a bidirectional view, bringing the versatility of the learned representation on both VL understanding and generation tasks.  

\vspace{-1.0em}
\paragraph{\textbf{Unimodal Event Encoding.}} The event encoding module aims to encode a video $\mathbf{V}$ into event features $\mathbf{E}$.
Inspired by the excellent long-term dependency modeling ability of Transformer~\cite{vaswani2017attention,zhu2020deformable}, we employ a transformer-based event detector to capture inter-frame relationships and propose event-level features. Specifically, we first use a pre-trained frame encoder (\eg. C3D~\cite{tran2015learning}) to extract frame-level features and refine them by a transformer encoder. Afterward, the transformer decoder uses $N$ learnable event queries to query frame features, during which each query gathers related foreground frames to form an event-level instance. We regard output features of event queries as the event-level representation $\mathbf{E}$, which stands for $N$ event proposals.

\vspace{-1.0em}
\paragraph{\textbf{Text-to-Event Grounding (TEG).}}
Given a set of sentence queries, TEG aims to ground the corresponding events from all predicted events $\mathbf{E}$ in parallel. 
By pulling the sentence query and its paired event together and pushing other events away, the cross-modal alignments are achieved, and the discriminability of visual features is enhanced. 
Given the features of an event set $\mathbf{E}=\{e_1, ..., e_N\}$ and sentence queries $\mathbf{Q}=\{q_1, ..., q_K\}$, we project them into a joint VL space and calculate the cross-modal cosine similarity matrix between the projected embeddings. 
We denote the cosine similarity matrix as $\boldsymbol{\omega} \in \mathbb{R}^{K \times N}$, where $\boldsymbol{\omega}(k, i)$ denotes the semantic-level similarity between $k$-th sentence and $i$-th event. After the label assignment (described in Sec.~\ref{sec:SAM}), we calculate the text-event contrastive loss: 
\vspace{-0.5em}
\begin{equation}
    \mathcal{L}_{\rm  teg} = - \sum_{k=1}^{K} T_{k}^{+} \log(\frac{\exp(\boldsymbol{\omega}(k, T_{k}^{+})/\tau)}{ \sum_{i=1}^{N}\exp(\boldsymbol{\omega}(k, i)/\tau) }),
\end{equation}
where $T_{k}^{+}$ is the assigned label for $k$-th sentence, $\tau$ is a temperature ratio.

For text encoding, we encode the input sentence set (paragraph) at different granularities by a hierarchical text encoder. Each sentence is first encoded by a RoBERTa encoder, followed by a self-attention layer to aggregate output tokens of all words to form a sentence embedding $\mathbf{Q}_{\rm sent}$. Finally, $\mathbf{Q}_{\rm sent}$ are concatenated with sentence-level position embedding, serving as the input of an additional self-attention layer for cross-sentence context modeling. 
Note that we construct two types of $\mathbf{Q}$, \ie, $\mathbf{Q}_{\rm ctx}$ and $\mathbf{Q}_{\rm sent}$ for sentence embedding with and without context modeling. The final loss of $\mathcal{L}_{\rm teg}$ is the summation of two TEG losses with $\mathbf{Q}_{\rm ctx}$ and $\mathbf{Q}_{\rm sent}$, respectively.
\vspace{-1.0em}
\paragraph{\textbf{Event-to-Text Generation (ETG).}} 
For capturing rich semantic features, we leverage ETG to predict ground-truth sentence annotations $\mathbf{Y}$, \ie, the sentences as well as their temporal locations. Toward this goal, two modules are attached upon $\mathbf{E}$, \ie, an LSTM-based lightweight text generator for generating sentences and an MLP-based localizer for predicting the time boundaries and confidences. The loss of ETG is calculated by $\mathcal{L}_{\rm etg} = \mathcal{L}_{\rm ce} + \mathcal{L}_{\rm loc}$, where $\mathcal{L}_{\rm ce}$ represents the cross-entropy loss between predicted words and ground-truth, $\mathcal{L}_{\rm loc}$ represents the summation of the boundary IoU loss and cross-entropy loss for confidence.

\subsection{Semantic-Aware Label Assignment}
\label{sec:SAM}
Label assignment tackles the problem of assigning a ground-truth instance to each predicted event as the learning goal. 
\textcolor{black}{
Previous DETR-based methods~\cite{carion2020end, kamath2021mdetr, wang2021end, lei2021detecting} consider the label assignment as an optimal matching problem with localization cost. Specifically, they use IoU cost and classification cost (\ie, the classification confidence on the ground-truth categories) as the matching cost, and then compute the optimal bipartite matching with the lowest cost.} We argue that both IoU cost and classification cost have their limitations. The IoU cost is easily affected by the unreliable, ambiguous boundary annotations, which are primarily caused by the subjective bias of annotators as indicated by~\cite{otani2020uncovering,huang2022video,chen2020refinement}. 
\textcolor{black}{As an untrimmed video usually has multiple events within the same action class, the classification cost at the coarse level is ineffective to distinguish the predictions with fine-grained differences. }

We propose a semantic-aware matching cost to circumvent the above dilemma. Specifically, a semantic cost is proposed, which utilizes the cross-modal similarity matrix $\boldsymbol{\omega}$ to measure the matching degree between predicted events and ground-truth sentences at the semantic level. Note that the similarity matrix is used for contrastive learning, making the similarity score have a favorable discriminability. Our final cost combines semantic cost with localization cost together. Specifically, the semantic cost between the $k$-th ground-truth element and $i$-th prediction element is measured by $-\boldsymbol{\omega}(k, i)$. We consider $\mathcal{L}_{\rm loc}$ as the localization cost, 
which measures  the matching degree across the temporal dimension. The final matching cost is denoted as:
\begin{equation}
    C = -\lambda\boldsymbol{\omega} + \mathcal{L}_{\rm loc},
    \label{eqn:sam}
\end{equation}
where $\lambda$ is the ratio of semantic cost. By performing the Hungarian matching algorithm~\cite{kuhn1955hungarian}, one-to-one matching is achieved. And the final loss is calculated by Eqn.~\ref{eqn:loss} with the assigned labels accordingly. 
\textcolor{black}{Compare with previous methods, we incorporate natural language into the matching cost, thereby increasing its sensitivity to subtle distinctions among predictions. Consequently, our approach exhibits improved robustness against ambiguous annotations.}

\section{Experiments}
\subsection{Experimental Setup}

\paragraph{Tasks.} We conduct experiments on four tasks, which could be divided into two categories: 1) Cross-modal generation tasks: dense video captioning (DVC) and video paragraph captioning (PC); 2) Cross-modal understanding tasks: single-/multi-sentence video grounding (SSVG/MSVG). For DVC and PC, we utilize the reconstructed sentences from the text generator and their time boundaries from the localizer as the prediction results. For SSVG/MSVG, we use the similarity matrix $\boldsymbol{\omega}$ to select the most similar event features for each sentence query, which is further decoded as a temporal boundary. Note that for SSVG, we use $\mathbf{Q}_{\rm sent}$ to calculate $\boldsymbol{\omega}$ since cross-sentence contexts are inaccessible by each sentence query.

\vspace{-1.0em}
\paragraph{Datasets.} The proposed framework is evaluated on four public datasets: ActivityNet Captions~\cite{krishna2017dense}, TACoS~\cite{regneri2013grounding}, YouMakeup~\cite{wang2019youmakeup} and YouCook2~\cite{zhou2018towards}. We follow the standard split utilized in previous work. More details about the datasets can be found in Sec.~\ref{supp_dataset}.

\vspace{-1.0em}
\paragraph{Implementation Details.}
We adopt the C3D~\cite{tran2015learning} pre-trained on Sports1M~\cite{karpathy2014large} to extract frame features for ActivityNet Captions and TACoS. For a fair comparison with state-of-the-art methods on ActivityNet Captions, we also evaluate our model based on TSP~\cite{wang2018temporal} features. For YouCook2, we adopt the TSN features provided by~\cite{zhou2018end}. For YouMakeup, we use the I3D features provided by~\cite{wang2019youmakeup}.

The transformer for event encoding is implemented as a Deformable Transformer~\cite{zhu2020deformable} with 2 encoder-decoder layers. We follow PDVC~\cite{wang2021end} to use LSTM-DSA as the text generator. The number of event queries $N$ is set to 30 for ActivityNet Captions and 100 for other datasets. For model training, we use AdamW optimizer with batch size 1 and learning rate of $5e^{-5}$ for ActivityNet Captions/YouCook2, and batch size 4 and learning rate of $1e^{-4}$ for YouMakeup/TACoS. The RoBERTa encoder is frozen for ActivityNet Captions/YouCook2 and trained with a learning rate of $1e^{-5}$ for TACoS/YouMakeup, respectively. The weight decay is set to $1e^{-4}$ for all datasets. The trade-off factor $\alpha$ is set to 0.05. The temperature $\tau$ is set to 0.1. The ratio of semantic cost $\lambda$ is set to 1.0. More details can be found in Sec.~\ref{supp_details}.  

\subsection{Comparison with State-of-the-art Methods}

\paragraph{\textbf{Dense Video Captioning.}} In Table ~\ref{table:SotaANET}, our method could surpass the state-of-the-art methods on most metrics on ActivityNet Captions, both with cross-entropy training and reinforcement training. Adopting TSP features could further boost the performance and establish the new state-of-the-art under these two training scenarios. Compared with PDVC, our framework exploits the bidirectional information flow between visual features and text, which increases the expressive ability of the event encoding network in two aspects: 1) TEG guides the event features towards a better coverage of salient video segments, advancing the generated captions to reveal a full video story; 2) label assignment strategy is enhanced by semantic cost, which could achieve a better assignment to link results than PDVC, especially for videos with noisy boundary annotations. In Table~\ref{table:SotaYC} and Table~\ref{tab:ym}, performance comparisons on YouCook2 and YouMakeup further verify the effectiveness of our method, which surpasses previous methods with a clear margin. We achieved 1st place in the MDVC@PIC Challenge 2022\footnote{Leaderboard at \url{https://codalab.lisn.upsaclay.fr/competitions/5102}}. 
\vspace{-1.0em}
\paragraph{\textbf{Video Paragraph Captioning.}} Generating a coherent paragraph requires a global perception of the foreground events and getting rid of the noisy segments. Table~\ref{table:SotaParaCap} shows the PC performance comparison on ActivityNet Captions.  Our method with  TSP features (RGB modality) could achieve similar performance with TDPC, the state-of-the-art paragraph captioner based on Transformer. We note that TDPC utilizes the combination of three feature extractors (ResNet-200~\cite{he2016deep} + I3D RGB + I3D Flow~\cite{carreira2017quo}), which can not be fairly compared with our method. Compared with the method relying on localization annotations during inference, our method could largely reduce the cost of manpower, meanwhile achieving comparable or better performance.

\begin{table}[t]
\caption{Dense captioning performance on ActivityNet Captions. 
}

\small
\vspace{-0.6em}
\renewcommand\arraystretch{0.9}
  \centering
     \makeatletter\def\@captype{table}\makeatother
     \setlength{\tabcolsep}{0.75 mm}{
\begin{tabular}{l|c|cccc}
    \toprule
    Method & Features & BLEU@4 & METEOR  & CIDEr & SODA(c) \\
    \midrule
    \multicolumn{6}{l}{\textbf{with cross-entropy training}} \\
    DCE~\cite{krishna2017dense} & C3D & 0.17 & 5.69 & 12.43& - \\
    TDA-CG~\cite{wang2018bidirectional} & C3D & 1.31 & 5.86 & 7.99 & - \\
    DVC~\cite{li2018jointly} & C3D & 0.73 & 6.93 & 12.61 & - \\
    Efficient~\cite{Suin2020efficient} & C3D & 1.35 & 6.21 & 13.82 & -  \\
    SDVC~\cite{Mun2019stream} &C3D & - & 6.92 & - & -\\
    ECHR~\cite{wang2020event} & C3D & 1.29 & 7.19 & 14.71 & 3.22 \\
    {PDVC}~\cite{wang2021end} & C3D & 1.65 & \textbf{7.50} & {25.87} & {5.26} \\
    UEDVC~\cite{zhang2022unifying} & C3D & 1.45 & 7.33 & \textbf{26.92} & 5.29 \\
    \rowcolor{Gray}
    {Ours} & C3D & \textbf{1.69} & {7.46} & {26.23} & \textbf{5.44} \\
    \midrule
    MT~\cite{zhou2018end} & TSN & 1.15 & 4.98 & 9.25 & - \\
    BMT~\cite{Iashin2020better} & I3D & 1.88 & 7.43 & 11.94& - \\
    {PDVC}~\cite{wang2021end}& TSP &{2.17} & {8.37} & {31.14} & {6.05} \\
    \rowcolor{Gray}
    {Ours} & TSP & \textbf{2.18} & \textbf{8.50} & \textbf{32.76} & \textbf{6.22} \\
    \midrule
    \multicolumn{6}{l}{\textbf{with reinforcement training}} \\
    SDVC~\cite{Mun2019stream} & C3D & 0.93 & 8.82 & \textbf{30.68}& -\\
    SGR~\cite{deng2021sketch} & C3D &\textbf{1.67} & {9.07} & 22.12 & - \\
    \rowcolor{Gray}
    {Ours} & C3D & 0.81 & {\textbf{9.13}} & 26.00 & \textbf{5.91} \\
    \midrule
    MFT~\cite{xiong2018move} & TSN & \textbf{1.24}& 7.08 & 21.00 & - \\
    SGR~\cite{deng2021sketch} & TSN & - & 9.37 & - & 5.29 \\
    \rowcolor{Gray}
    {Ours} & TSP & 1.11 & \textbf{10.03} & \textbf{33.33} & \textbf{7.11} \\
    \bottomrule
\end{tabular}}
\label{table:SotaANET}
\vspace{-1.0 em}
\end{table}

\begin{table}[t]
\caption{Dense captioning on YouCook2 with TSN features.}
\vspace{-0.6em}
\small
\renewcommand\arraystretch{0.9}
       \centering
        \makeatletter\def\@captype{table}\makeatother
        \setlength{\tabcolsep}{2.2 mm}
        {
        \begin{tabular}{lcccc}
            \toprule
            \multirow{1}{*}{Method}  & BLEU@4 & \multirow{1}{*}{METEOR} & CIDEr & \multirow{1}{*}{SODA(c)} \\
            \midrule
            MT~\cite{zhou2018end} & 0.30 & 3.18  & 6.10 & - \\
            ECHR~\cite{wang2020event} & - & 3.82 & - & -\\
            SGR~\cite{deng2021sketch} & - & 4.35 & - & - \\
            {PDVC~\cite{wang2021end}} & 0.80 & {4.74}  & 22.71 & {4.42} \\
            \rowcolor{Gray}          
            {{Ours}} & \textbf{1.04} & \textbf{5.01} & \textbf{26.52} & \textbf{4.91} \\            
            \bottomrule
        \end{tabular}}
        \label{table:SotaYC}
        \vspace{-1.0em}
\end{table}

\begin{table}[]
\caption{Paragraph captioning on ActivityNet Captions. Anno. means needing boundary annotations during inference. $*$ means extra input modalities (\eg, flow, object detection) except RGB. 
}

\renewcommand\arraystretch{0.9}
\small
\vspace{-0.6em}
\centering
        \makeatletter\def\@captype{table}\makeatother
        \setlength{\tabcolsep}{1.35 mm}{
\begin{tabular}{lcccc}
    \toprule
    \multirow{1}{*}{Method}  & \multirow{1}{*}{Anno.} & \multirow{1}{*}{BLEU@4} & \multirow{1}{*}{METEOR}  & \multirow{1}{*}{CIDEr} \\
    \midrule
    MFT$^{*}$~\cite{xiong2018move} & $\times$ & {10.29} & 14.73  & 19.12 \\
    {PDVC}~\cite{wang2021end} & $\times$ & 10.46 & \textbf{16.42} & 20.91 \\
    {TDPC$^{*}$ (w/o RL)}~\cite{song2021towards} & $\times$ & \textbf{11.74} & 15.64 & {\textbf{26.55}} \\
    \rowcolor{Gray}
    {Ours} & $\times$ & \underline{11.70} & \underline{16.35} & \underline{26.03} \\
    \midrule
    {\color{gray}HSE~\cite{zhang2018cross}} & {\color{gray}\checkmark} &{\color{gray}9.84} & {\color{gray}13.78} & {\color{gray}18.78} \\
    {\color{gray}MART$^{*}$~\cite{lei2020mart}}  & {\color{gray}\checkmark} & {\color{gray}10.33} & {\color{gray}15.68} & {\color{gray}23.42} \\
    {\color{gray}GVD$^{*}$~\cite{zhou2019grounded}} & {\color{gray}\checkmark} & {\color{gray}11.04} & {\color{gray}15.71} & {\color{gray}21.95} \\
    {\color{gray}AdvInf$^{*}$~\cite{park2019adversarial}} & {\color{gray}\checkmark} & {\color{gray}10.04} & {\color{gray}16.60} & {\color{gray}20.97} \\
    {\color{gray}COOT$^{*}$~\cite{ging2020coot}} & {\color{gray}\checkmark} & {\color{gray}10.85} & {\color{gray}15.99} & {\color{gray}{28.19}} \\
    \bottomrule
\end{tabular}}
\label{table:SotaParaCap}
\vspace{-1.0em}
\end{table}

\begin{table}[]
\caption{Dense captioning and multi-sentence video grounding performance on YouMakeup. For the validation set, all approaches utilize I3D features, while for the test set, $\ddagger$ means using 2D and 3D features, and ${\star}$ means using model ensembling. }
\small
\vspace{-0.6em}
\renewcommand\arraystretch{0.9}
\centering
        \makeatletter\def\@captype{table}\makeatother
        \setlength{\tabcolsep}{1.0 mm}{
\begin{tabular}{l|cc|ccc}
\toprule
    \multirow{2}{*}{Method} & \multicolumn{2}{c|}{{Dense Captioning}} &\multicolumn{3}{c}{{Grounding}} \\
 & METEOR & SODA(c) & IoU0.5   & IoU0.7   & mIoU \\ 
 \midrule
  \multicolumn{6}{l}{\textbf{Validation Set}} \\
PDVC~\cite{wang2021end} & 21.86 & 17.12 & - & - & -  \\
MMN~\cite{wang2021negative} & - & - & 35.18 & 20.08 & -  \\
\rowcolor{Gray}
{Ours} & \textbf{22.28} & \textbf{18.93} & \textbf{55.70} & \textbf{38.24} & \textbf{51.02}  \\
\midrule
    \multicolumn{6}{l}{\textbf{Leaderboard on Evaluation Server (Test set)}} \\
Lu et.al$^{\ddagger}$~\cite{mdvc2nd} & 23.17 & - & - & - & - \\    
Shu et.al$^{\ddagger \star}$~\cite{mtvg3rd} & - & - & 56.01 & 38.44 & - \\
Zhang et.al$^{\ddagger \star}$~\cite{mtvg2nd} & - & - & \textbf{62.50} & 42.12 & - \\
\rowcolor{Gray}
{Ours$^{\ddagger \star}$} & \textbf{37.58} & - & 61.56 & \textbf{46.18} & - \\    
\bottomrule
\end{tabular}
}
\label{tab:ym}
\vspace{-0.2em}
\end{table}

\vspace{-1.0em}
\paragraph{\textbf{Single-Sentence Video Grounding.}}

Table~\ref{tab:ssvg} shows the SSVG performance comparison on ActivityNet Captions and TACoS. The proposed method achieves state-of-the-art performance among metric learning methods and competitive performance with non-metric learning
methods. This verifies that the learned event representations are semantically rich enough to make them distinguishable from each other and are also temporally sensitive to differentiate themselves from many background segments. Compared with non-metric learning methods, our method could reuse the visual feature when querying the same video with different sentences, which is more efficient during training and inference. MMN is the state-of-the-art metric learning grounding method that adopts a contrastive learning loss similar to ours. However, its label assignment is merely based on localization cost, which lacks semantic similarity and thus may suffer from incorrect boundary annotations. We surpass MMN on TACoS  with a clear margin.

\begin{table}[]
\caption{Single-sentence video grounding on ActivityNet Captions and TACoS.  All methods are based on C3D features without special indication.}
\vspace{-0.5em}
\renewcommand\arraystretch{1.0}
\centering
\small
        \makeatletter\def\@captype{table}\makeatother
        \setlength{\tabcolsep}{0.8 mm}{
\begin{tabular}{l|ccc|ccc}
\toprule
    \multirow{2}{*}{Method} &\multicolumn{3}{c|}{{ActivityNet Captions}} & \multicolumn{3}{c}{{TACoS}} \\ & IoU0.5   & IoU0.7   & mIoU & IoU0.3   & IoU0.5 & mIoU  \\
\midrule
\multicolumn{7}{l}{\textbf{Non-metric learning based}} \\

CTRL \cite{gao2017tall} & 29.01 & 10.34 & 20.54  & 18.32 & 13.30 & 11.98\\
TGN \cite{chen2018temporally} & 27.93 & 11.86 & 29.17 & 21.77 & 18.90 & 17.93\\
QSPN \cite{xu2019multilevel} & 33.26 & 13.43 & - & 20.15 & 15.23 & -  \\
ABLR \cite{yuan2019find} & 36.79 & - & 36.99 & 19.50 & 9.40  & 13.40 \\
DRN \cite{zeng2020dense} & 45.45 & 24.36 & -  & - & 23.17 & -   \\
2D-TAN \cite{zhang2020learning}& 44.51 & 26.54 & - & 37.29 & 25.32 & - \\
VSLNet \cite{zhang2020span} & 43.22 & 26.16 & \textbf{43.19} & 29.61 & 24.27 & 24.11 \\
CPNet \cite{li2021proposal} & 40.56 & 21.63 & 40.65 & 42.61 & 28.29 & \textbf{28.69} \\
BPNet\cite{xiao2021boundary} & 42.07 & 24.69 & 42.11 & 25.96 & 20.96 & 19.53 \\
CBLN\cite{liu2021context} & 48.12 & 27.60 & - & 38.98 & 27.65 & -   \\
MATN\cite{zhang2021multi} & 48.02 & \underline{31.78} & - & \textbf{48.79} & \textbf{37.57} & - \\
MGSL-Net\cite{liu2022memory} & \underline{51.87} & 31.42 & - & 42.54 & 32.27 & - \\
D-TSG~\cite{liu2022reducing} & \textbf{54.29} & \textbf{33.64} & - & \underline{46.32} & \underline{35.91} & - \\
\midrule
\multicolumn{7}{l}{\textbf{Metric learning based}} \\
MCN \cite{anne2017localizing}   & 21.36 & 6.43  & 15.83  & -     & 5.58  & -  \\
MMN\cite{wang2021negative} & 48.59 & \underline{29.26} & - & \underline{39.24} & \underline{26.17} & - \\
\rowcolor{Gray}
{Ours (C3D) } & \underline{48.93} & 27.16 & \underline{46.38} & \textbf{45.92} & \
\textbf{34.57} & \textbf{32.48}\\
\rowcolor{Gray}
{\color{black}
{Ours (TSP)}} & \textbf{49.18} & \textbf{29.69} & \textbf{46.83} & - & - & - \\

\bottomrule
\end{tabular}}
\label{tab:ssvg}
\vspace{-1.0em}
\end{table}

\vspace{-1.0em}
\paragraph{\textbf{Multi-Sentence Video Grounding.}}
MSVG performs a set-level prediction, which takes all text queries as input and localizes their target events simultaneously. 
As shown in Table~\ref{tab:msvg} and Table~\ref{tab:ym}, 
we establish a new state-of-the-art on TACoS and Youmakeup, meanwhile achieve competitive performance on ActivityNet Captions. We also achieved 1st place in the MTVG@PIC Challenge 2022\footnote{Leaderboard at \url{https://codalab.lisn.upsaclay.fr/competitions/5244}}. We conjecture the advantage of our method comes from 
the one-to-one matching strategy with a semantic-aware cost at the set level. Our method guarantees the coherence of the matching results, which could suppress the hard negative segments that have a high IoU but low semantic similarity.

\begin{table}[]
\caption{Multi-sentence video grounding performance on ActivityNet Captions and TACoS. All methods are based on C3D features without special indication.}
\vspace{-0.5em}
\renewcommand\arraystretch{0.9}
\centering
\small
        \makeatletter\def\@captype{table}\makeatother
        \setlength{\tabcolsep}{1.0 mm}{
\begin{tabular}{l|ccc|ccc}
\toprule
    \multirow{2}{*}{Method} &\multicolumn{3}{c|}{{ActivityNet Captions}} & \multicolumn{3}{c}{{TACoS}} \\
 & IoU0.5   & IoU0.7   & mIoU & IoU0.3   & IoU0.5   & mIoU  \\ \midrule

BS \cite{bao2021dense} & 46.43 & 27.12 & - & 38.14 & 25.72 & -   \\
3D-TPN \cite{zhang2020learning} & 51.49 & 30.92 & - & 40.31 & 26.54 & -   \\
DepNet \cite{bao2021dense} & 55.91 & 33.46 & - & 41.34 & 27.16 & -   \\
SVPTR \cite{jiang2022semi} & \textbf{61.70} & \underline{38.36} & \textbf{55.91} & \underline{47.89} & \underline{28.22} & \underline{31.42} \\
\rowcolor{Gray}
{Ours (C3D)} & 59.08 & 37.47 & 54.56 & \textbf{48.29} & \textbf{36.07} & \textbf{34.29} \\
\rowcolor{Gray}

{\color{black}
{Ours (TSP)}} & \underline{60.67} & \textbf{38.55} & \underline{55.40} & - & - & - \\

\bottomrule
\end{tabular}}
\label{tab:msvg}
\vspace{-0.2em}
\end{table}

\subsection{Ablation Studies} \label{sec-abl}
We study the key design choices of the proposed framework on ActivityNet Captions, including the bidirectional pretext tasks, matching cost in label assignment, and the choices of text encoding. 
\vspace{-1.0 em}
 \paragraph{\textbf{Bidirectional Pretext Tasks.}} After training, we study the expressive ability of the event-level features learned by bidirectional pretext tasks.  Specifically, a caption head (vanilla LSTM) and an FC layer are attached to the event features to evaluate the caption prediction ability and classification ability, respectively. By the linear probing performance of caption prediction and temporal action localization, we investigate whether the bidirectional prediction learns better representations than unidirectional prediction. 
As shown in Table~\ref{tab:bp}, our final model with bidirectional prediction performs better than models trained with merely ETG or TEG tasks in terms of several captioning metrics, and the average mAP score of temporal action localization also shows the same trend, which verifies the effectiveness of the proposed bidirectional pretext tasks.

\begin{table}[]
\caption{Ablation studies of bidirectional pretext tasks. The performance of temporal action localization is evaluated on ActivityNet 1.3~\cite{caba2015activitynet}, which shares the same videos with ActivityNet Captions. For ``w/o ETG", we remove the caption generator, and for ``w/o TEG", we remove the text encoding branch. }
\vspace{-0.5em}
\renewcommand\arraystretch{0.9}
\centering
    \small
    \makeatletter\def\@captype{table}\makeatother
        \setlength{\tabcolsep}{1.15 mm}{
\begin{tabular}{l|ccc|c}
\toprule
    \multirow{2}{*}{Method} & \multicolumn{3}{c|}{{Captioning}} & Localization  \\
  & METEOR(PC) & METEOR & SODA(c) & Avg mAP \\ 
 \midrule
w/o TEG & 14.18 & 7.50 & 4.65 & 18.83 \\
w/o ETG & 13.66 & 6.84 & 4.80 & 12.40  \\
\rowcolor{Gray}
{Full} & \textbf{15.73} & \textbf{8.03} & \textbf{5.68} & \textbf{22.28} \\
\bottomrule

\end{tabular}}
\label{tab:bp}
\vspace{-1.0em}
\end{table}

 \begin{table}[]
\caption{Comparison of different semantic-aware costs.}
\small
\vspace{-0.5em}
\renewcommand\arraystretch{0.9}
\centering
        \makeatletter\def\@captype{table}\makeatother
        \setlength{\tabcolsep}{0.75 mm}{
\begin{tabular}{l|ccc|cc}
\toprule
    \multirow{2}{*}{Method} &\multicolumn{3}{c|}{{Grounding}} & \multicolumn{2}{c}{{Dense Captioning}} \\
 & IoU0.5   & IoU0.7   & mIoU & METEOR & SODA(c) \\ 
 \midrule
w/o semantic cost & 58.30 & 36.82 & 53.96 & 8.70 & 5.63  \\
caption cost & 58.14 & 37.19 & 54.12 & \textbf{8.84} & 5.63  \\
\rowcolor{Gray}
{contrastive cost} & \textbf{60.67} & \textbf{38.55} & \textbf{55.40}  & 8.50 & \textbf{6.22} \\
\bottomrule
\end{tabular}
}
\label{tab:labelassign}
\vspace{-1.0em}
\end{table}

\begin{figure*}[]
    \centering
    \includegraphics[width=0.95\textwidth]{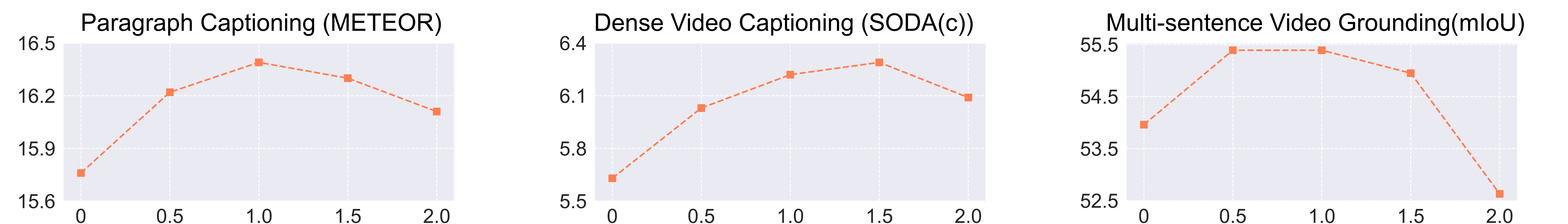}
         \vspace{-0.7 em}
    \caption{The performance on three tasks by varying contrastive cost ratios.}
    \label{fig:clratio}
    \vspace{-1.0em}
\end{figure*}

\vspace{-1.0em}
\paragraph{\textbf{Label Assignment.}} We test two types of cost to measure the semantic distance between an event and a sentence: 1) Contrastive cost; 2) Caption cost, i.e., the negative log probability of the target sentence produced by the caption head. Table~\ref{tab:labelassign} shows the ablation of semantic costs in label assignment. Only adopting localization cost can not capture the high-level similarity between events and sentences, resulting in inferior performance. By using caption cost, the matching algorithm focus on event proposals that could generate the target sentence with a high probability. However, we found that the joint probability of generated sentences is affected by the sentence length and the portion of common words. For example, short sentences containing common words usually have a lower matching cost with all event segments. We can see from the table that incorporating caption cost does not show a clear improvement. Our model with contrastive cost could achieve a clear performance gain on most metrics. Although there is a performance drop on METEOR, we achieve a substantial improvement gain on SODA(c). Considering that SODA(c) measure a set of captions together by considering their temporal structure, we conclude that the proposed contrastive cost could increase the coherence of generated captions and the accuracy of grounding results. 

Fig.~\ref{fig:clratio} shows the performance comparison by varying semantic cost ratio ${\lambda}$. From the sub-figures, we can see that after modulating the cost ratio, the listed metrics across four tasks could achieve performance gain over ${\lambda}=0$. As ${\lambda}$ becomes larger than 1.5, the performance on most tasks drops. It is reasonable since too large ${\lambda}$ will make the total matching cost overwhelmed by semantic similarity and neglects the localization cost, leading to wrong matching when semantically-similar events occur in different positions of the video.

\begin{table}[]
\caption{Ablation studies of text encoding.}
\vspace{-0.5em}
\small
\renewcommand\arraystretch{1.0}
\centering
        \makeatletter\def\@captype{table}\makeatother
        \setlength{\tabcolsep}{0.65 mm}{
\begin{tabular}{l|cc|cc}
\toprule
     \multirow{2}{*}{Settings} &\multicolumn{2}{c|}{{Grounding}} & \multicolumn{2}{c}{{Dense Captioning}} \\
& IoU0.7   & mIoU & METEOR & SODA(c) \\ 
\midrule
CLS& 22.17 & 42.14 & 8.27 & 6.20 \\
Mean& 28.02 & 46.74 & 8.30 & \textbf{6.24} \\
Att & 29.69 & 46.83 & 8.42 & 6.13 \\
Att + Cross-Sent Att& 32.11 & 48.88 & 8.34 & 6.17 \\
\rowcolor{Gray}
Att + Cross-Sent Att + PE& \textbf{38.55} & \textbf{55.40} & \textbf{8.50} & 6.22\\ 
\bottomrule
\end{tabular}}
\label{tab:te}
\vspace{-1.0em}
\end{table}

\vspace{-1.0em}
\paragraph{\textbf{Text Encoding}}

Table~\ref{tab:te} shows the ablation of text encoding. We first study the ablations of sentence encoding. ``CLS" considers the [CLS] tokens produced by RoBERTa as the sentence representation. ``Mean" means we obtain the sentence representation by calculating the mean-pooled features of all output tokens of RoBERTa. Our model with attention aggregation (``Att") could achieve superior performance over these two variants on grounding and comparable performance on dense captioning. Additionally, we found that position embedding (PE) in context modeling plays an essential role in boosting the grounding performance. We conclude that the order of sequence is crucial for localization since essential sentence orders are typically consistent with the temporal order of the target events.

\begin{figure}[]
    \centering
    \includegraphics[width=0.5\textwidth]{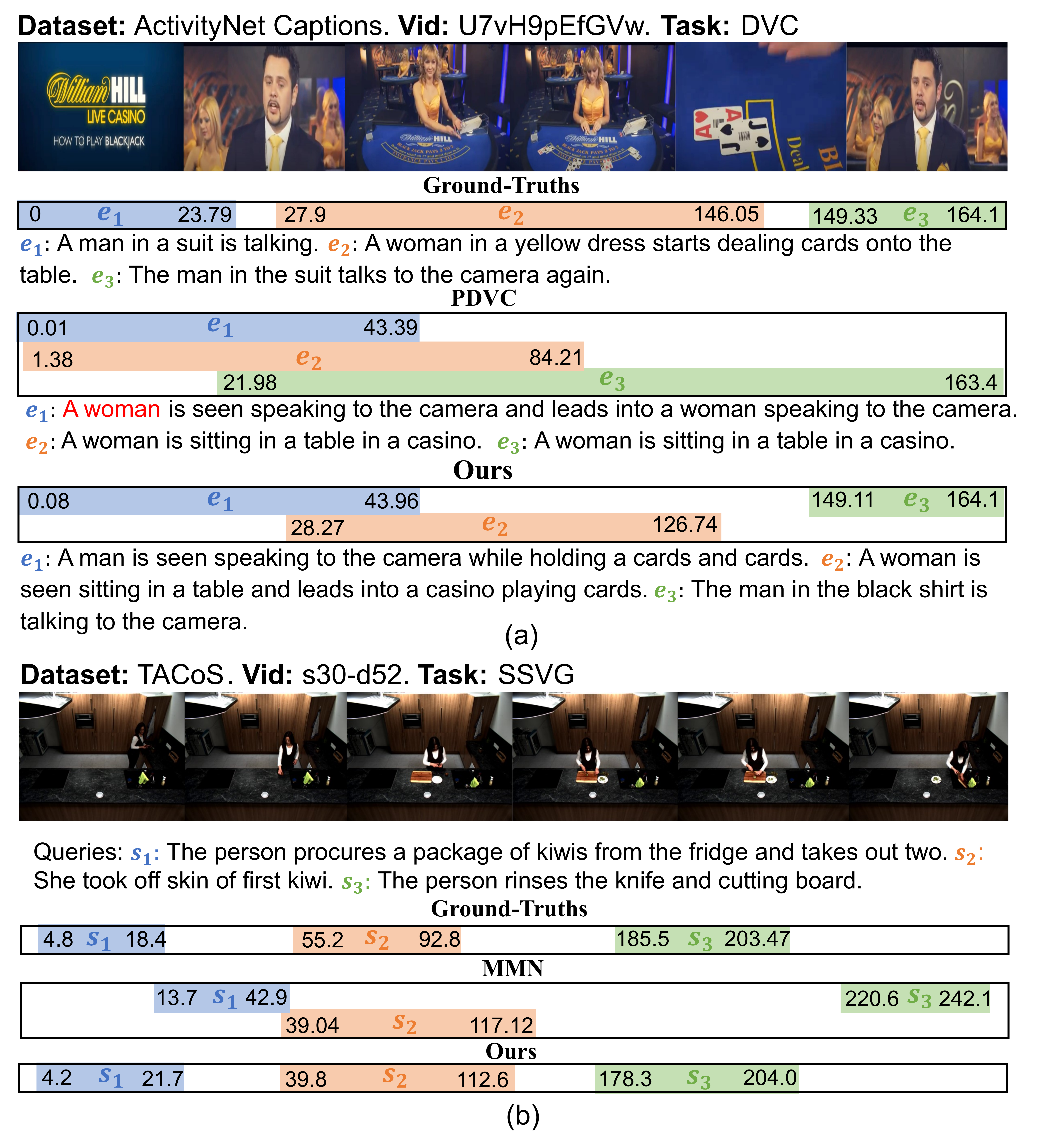}
         \vspace{-2.2em}
    \caption{Visualization of video examples.}
    \label{fig:qa}
    \vspace{-1.5em}
\end{figure}

\subsection{Qualitative Analysis}
To illustrate the quality of learned event-level features, we visualize two examples of different tasks and compare them with previous state-of-the-art methods. As shown in Fig.~\ref{fig:qa}, the proposed method could surpass previous methods in both the quality of decoded sentences and the accuracy of predicted timestamps. Note that distinguishing semantically similar events is a challenging problem for untrimmed video learning. In Fig.~\ref{fig:qa}(a), PDVC mixes up multiple events and recognizes the wrong subject (``woman'' instead of ``man''). Compared with PDVC, our method benefits from the text supervision in TEG so that the model can understand fine-grained differences between events and preserve more accurate semantic information in event features.

\section{Conclusion}
This paper proposes a joint video-language learning framework to learn event-level representation from untrimmed videos. We present two pretext tasks, text-to-event grounding (TEG) and event-to-text generation (ETG), to learn discriminative and semantically-rich event features, which view  the exploitation of cross-modal interactions from the untrimmed video as solving a bidirectional set prediction problem. A novel semantic-aware matching cost is also proposed to mitigate the ambiguity of localization cost in label assignment. Extensive experiments demonstrate the effectiveness of the proposed method, and we surpass state-of-the-art methods on multiple cross-modal tasks.



{\small
\bibliographystyle{ieee_fullname}
\bibliography{egbib}
}

\renewcommand\thesection{\Alph{section}}
\renewcommand\thefigure{A\arabic{figure}}
\renewcommand\thefigure{A\arabic{figure}}
\newpage
\setcounter{section}{0}
\setcounter{figure}{0}
\section{Supplementary Materials}
\subsection{Datasets}
\label{supp_dataset}
\paragraph{\textbf{ActivityNet Captions.}~\cite{krishna2017dense}} It collects untrimmed videos covering diverse daily human activities from YouTube. The official dataset provider split 10,009, 4,917, and 5,044 videos for training, validation, and testing. On average, each video lasts 120 seconds and is annotated by 3.65 event-sentence pairs. We use ActivityNet Captions for the tasks of single-/multi-sentence video grounding, dense video captioning, and video paragraph captioning. Note that the validation set has two independent annotations, \ie, \textit{val1} and \textit{val2}. For video grounding tasks, we follow~\cite{zhang2019cross} to use \textit{val1} for validation and \textit{val2} for testing. For video paragraph captioning, we follow~\cite{lei2020mart, zhou2019grounded} to validate our model on the \textit{ae-val} set and report the testing results on the \textit{ae-test} split. For dense video captioning, we report overall results on \textit{val1} and \textit{val2}.

\vspace{-1.0 em}
\paragraph{\textbf{TACoS.}~\cite{regneri2013grounding}} This dataset contains 127 cooking videos from MPII-Cooking dataset~\cite{rohrbach2012script}. We follow the standard split~\cite{gao2017tall}, assigning 10,146, 4,589, and 4,083 event-sentence pairs of 75, 27, and 25 videos for training, validation, and testing, respectively.

\vspace{-1.0 em}
\paragraph{\textbf{YouCook2.}~\cite{zhou2018towards}} It consists of 2,000 long untrimmed videos covering 89 cooking recipes. The videos have an average duration of 5.25 minutes and 7.7 annotated event-sentence pairs. We follow the standard split, using 1,333, 457, and 210 videos for training, validation, and testing. Since the annotation of the test set is hold-out, we follow~\cite{zhou2018end} to report evaluation results on the validation set.

\vspace{-1.0 em}
\paragraph{\textbf{YouMakeup.}}~\cite{wang2019youmakeup} It contains 2800 makeup videos collected from YouTube. The total length of all videos is about 421 hours, and there are 30626 annotated event-sentence pairs. We follow the official split using 1680, 280, and 840 videos for training, validation, and testing, respectively. We report our evaluation result on the validation set since the test set is hold-out.     

\subsection{Further Implementation Details}
\label{supp_details}
\paragraph{\textbf{Training.}} We follow other DETR-style approaches~\cite{zhu2020deformable, wang2021end} to set the ratio of boundary IoU loss and confidence cross-entropy loss to 2 and 1, respectively. To alleviate the repeated prediction phenomenon caused by excessive event features in dense video captioning, we follow~\cite{wang2021end} to additionally attach an event counter to the event encoding module to predict the event count. The counter is trained by cross-entropy loss between the predicted count distribution and the ground-truth count number.

For TACoS, we adopt a similar training and testing strategy with~\cite{bao2021dense} during training and testing. We divide the annotated paragraph of each video into multiple sub-paragraphs by temporal order. Each sub-paragraph has eight sentences. If the paragraph has less than eight sentences, we directly set the whole paragraph as a sample. 
For the other datasets, we directly feed the whole paragraph during training and testing.

\paragraph{\textbf{Reinforcement Learning for DVC.}}
For the dense video captioning task in Table.~{1}, we perform an additional training stage, i.e., reinforcement learning (RL)~\cite{rennie2017self}, to fairly compare with other approaches. We use METEOR and CIDEr as the rewards. To improve the stability during RL training, we increase the number of data samples using a data augmentation method, \ie, \textit{random cropping}. Given a video, we randomly crop a span with a temporal length $r_{\rm crop}\times L$, where $L$ is the video duration and $r_{\rm crop}\sim U(0.5, 1)$ is a ratio to control the span length. $U(a,b)$ denotes the uniform distribution over the range $[a,b]$. For each video, we repeatedly sample $H=256$ spans to form a batch of samples for RL training.

\begin{table*}[]
\caption{Ablation studies of text encoding on ActivityNet Captions. The first line represents our default setting.}
\vspace{-1.0em}
\renewcommand\arraystretch{1.0}
\centering
        \makeatletter\def\@captype{table}\makeatother
        \resizebox{0.8\linewidth}{!}{
\begin{tabular}{l|cccc|cccc}
\toprule
    \multirow{2}{*}{Word encoder} &\multicolumn{4}{c|}{{Multi-Sentence Video Grounding}} & \multicolumn{4}{c}{{Dense Video Captioning}} \\
 & IoU0.3   & IoU0.5   & IoU0.7   & mIoU & BLEU@4 & METEOR & CIDEr & SODA(c) \\ 
\midrule
\rowcolor{Gray}
RoBERTa & {76.64} & {60.67} & \textbf{38.55} & \textbf{55.40} & \textbf{2.18} & \textbf{8.50} & \textbf{32.76} &\textbf{6.22}\\ 
BERT &\textbf{76.65} & \textbf{61.08} & 38.25 & 55.39 & 2.09 & 8.38 & 30.34 & 6.13 \\
DistilBERT & 76.35 & 60.45 & 38.06 & 55.35 & 2.08 & 8.32 & 30.67 & 6.11 \\
Vanilla & 75.65 & 58.70 & 37.73 & 54.71 & 2.13 & 8.31 & 31.57 & 6.14\\
\bottomrule
\end{tabular}}
\label{tab:supte}
\vspace{-1.0em}
\end{table*}

\paragraph{\textbf{Bidirectional Pretext Tasks.}} Regarding the ablation study of the bidirectional prediction in Sec.~$4.3$, we first pretrain the proposed framework, then two types of probing layers are attached to study the quality of learned event features. For pretraining, we set the training batch size to 4. During probing, only attached layers are trained, and all other layers are frozen. We study the caption prediction ability on ActivityNet Captions and classification ability on ActivityNet 1.3~\cite{caba2015activitynet}, respectively. For temporal action localization, we follow the official split~\cite{caba2015activitynet} and report the mean average precision on the validation set.

\subsection{Analysis and Visualization}
\label{supp_analysis}
\begin{figure}[]
    \centering
    \includegraphics[width=0.45\textwidth]{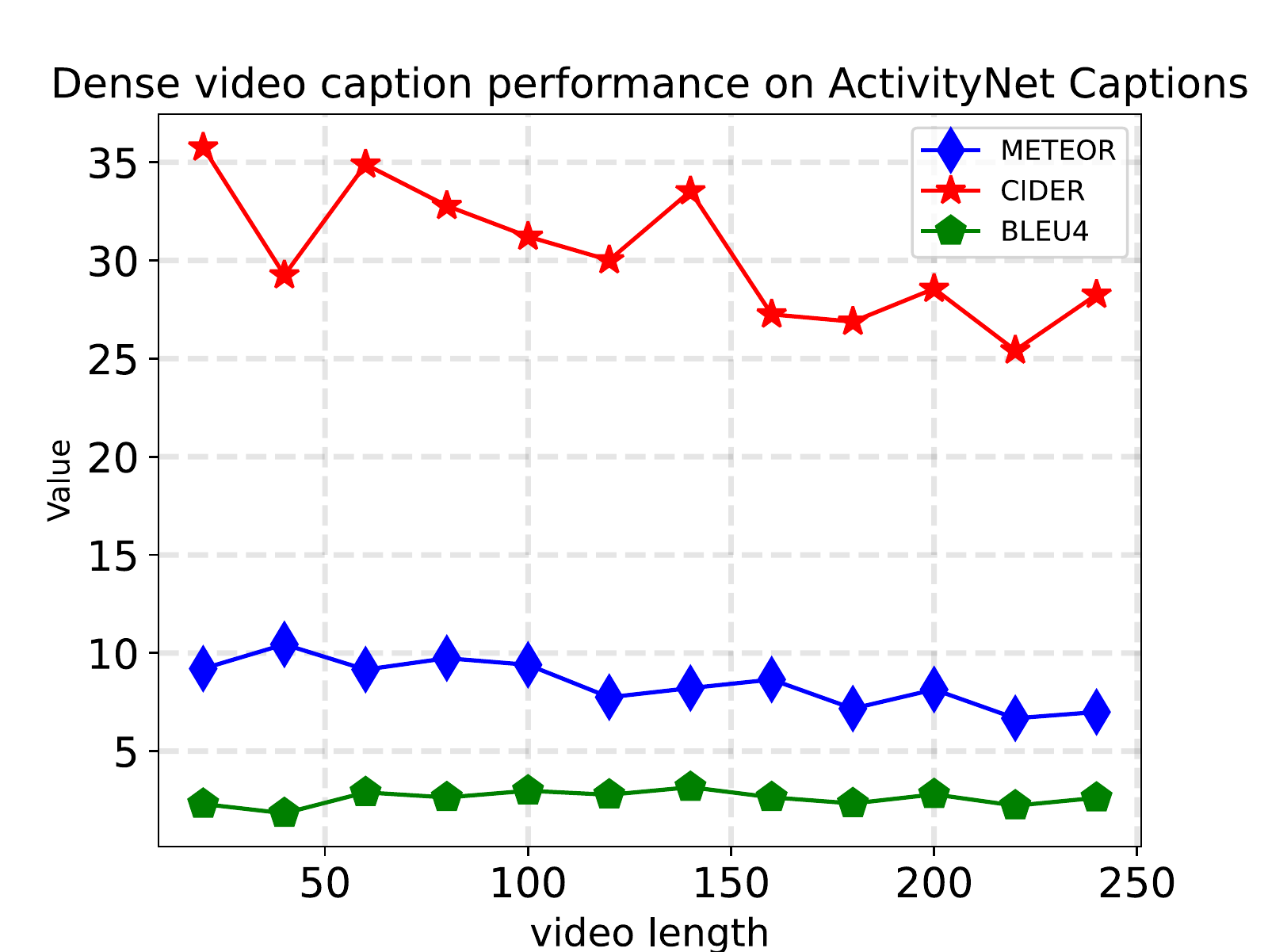}
    \caption{The dense video captioning performance on  videos with different video lengths.}
    \label{fig:vidlength}
    \vspace{-1.0em}
\end{figure}

\paragraph{\textbf{Text Encoding.}} We analyze the choice of the word encoder. As shown in Table~\ref{tab:supte}, the pretrained language models, \ie, BERT, DistilBERT, and RoBERTa work slightly better than the vanilla Transformer model without pretraining. We conclude that large-scale pretraining on general text corpus could enhance the text encoding process and then help to learn a better event-level representation that improves the grounding and dense captioning performance. 

\begin{figure}[]
    \centering
    \includegraphics[width=0.45\textwidth]{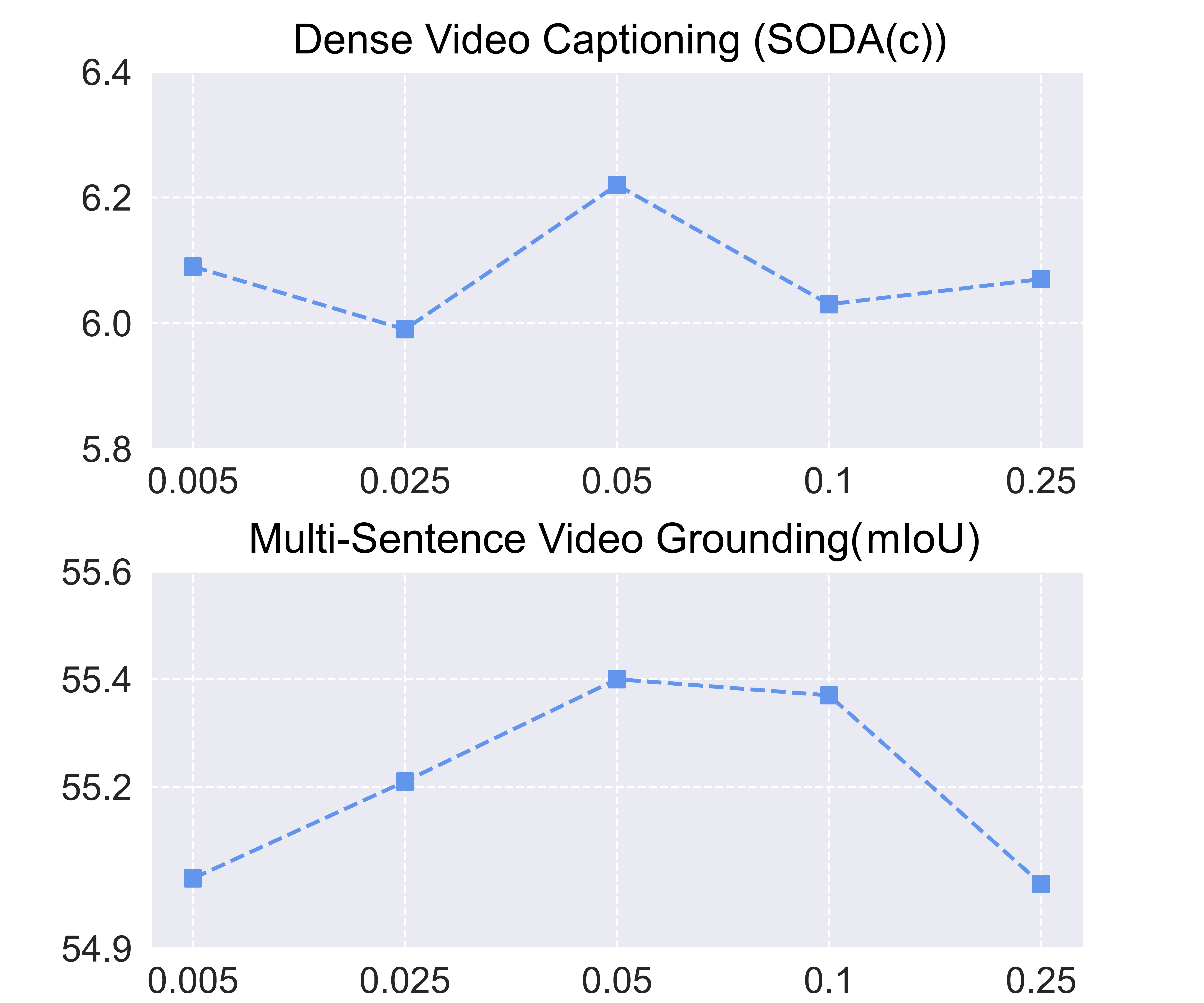}
    \caption{Performance comparison by varying loss ratio $\alpha$.}
    \label{fig:lossratio}
\end{figure}

\begin{figure}[]
    \centering
    \includegraphics[width=0.45\textwidth]{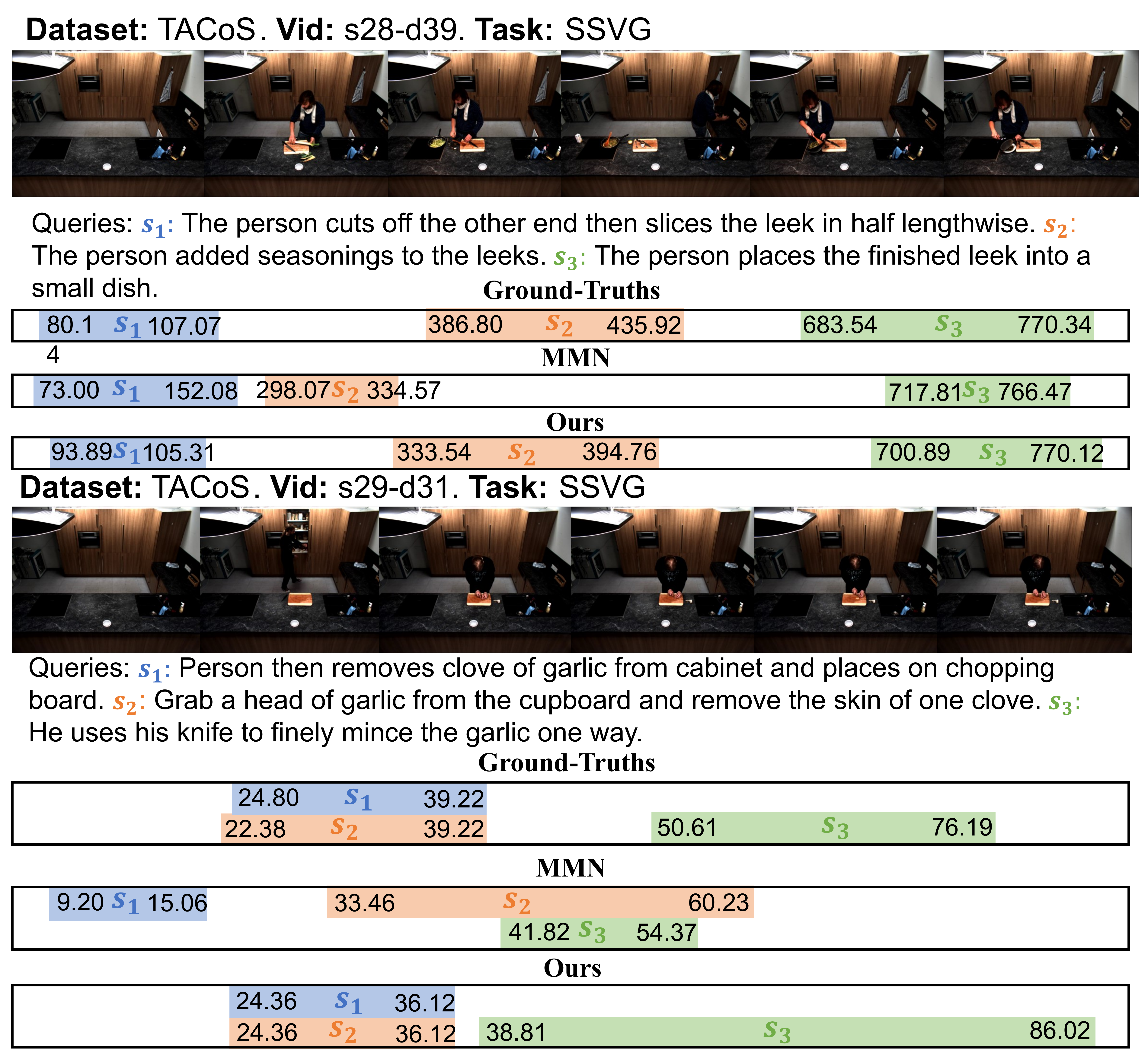}
         \vspace{-0.8em}
    \caption{Visualization of SSVG examples.}
    \label{fig:ssvgexamples}
\end{figure}

\begin{figure}[]
    \vspace{-0.8em}
    \centering
    \includegraphics[width=0.45\textwidth]{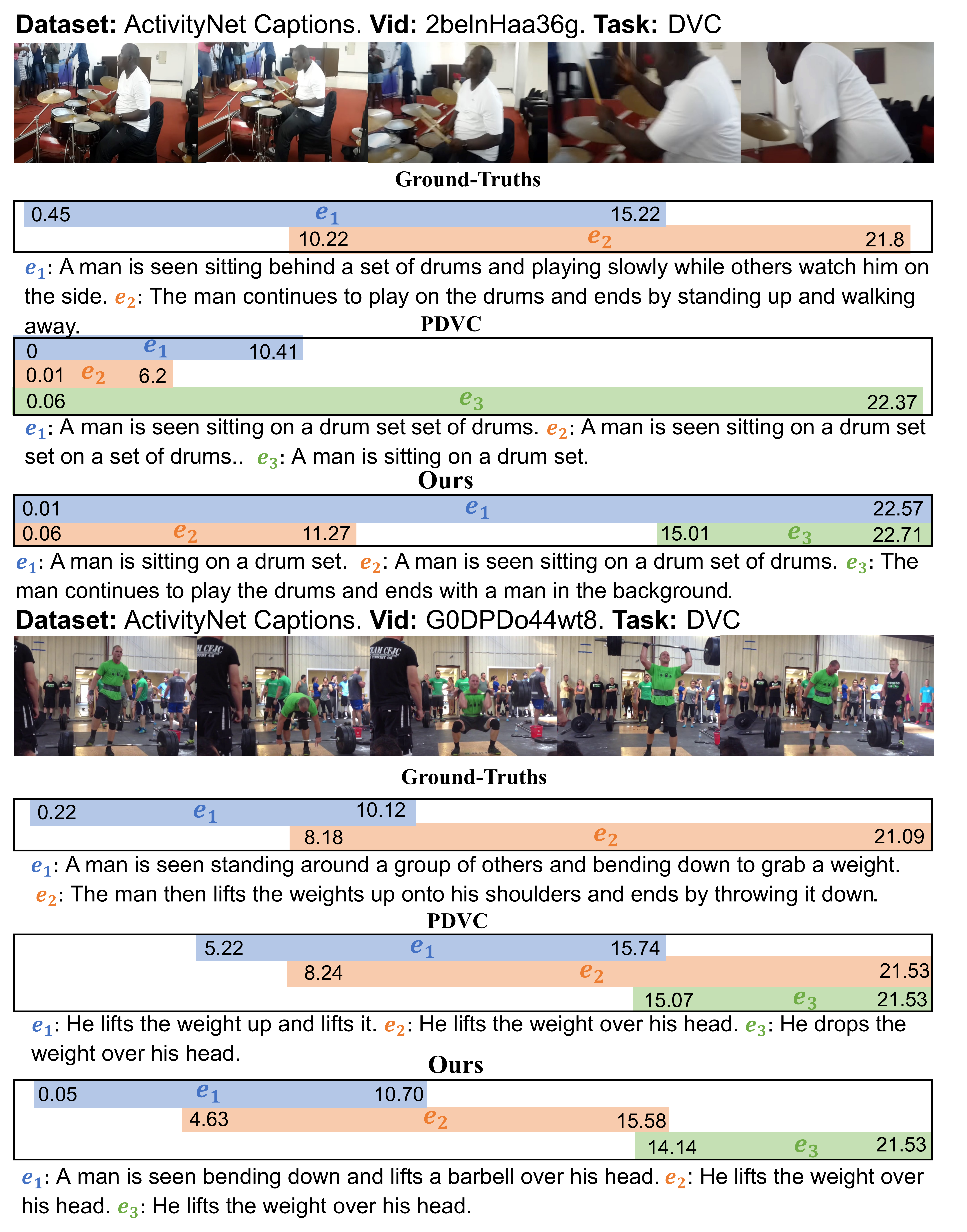}
         \vspace{-0.9em}
    \caption{Visualization of DVC examples.}
    \label{fig:dvcexamples}
\end{figure}

\vspace{-1.0em}
\paragraph{\textbf{Video Length.}}
Fig.~\ref{fig:vidlength} shows that the performance of our method on dense video captioning is affected by the video duration. For longer videos, the performance tends to be low, possibly caused by the large variance and sophisticated visual relationships in long videos. 
\vspace{-1.0em}
\paragraph{\textbf{Loss Ratio.}} We conduct an ablation study of $\alpha$ (the loss ratio of $\mathcal{L}_{\rm teg}$) on ActivityNet Captions. As shown in Fig.~\ref{fig:lossratio}, $\alpha=0.05$ achieves the best overall performance on MSVG in terms of mIoU and DVC in terms of SODA(c). A lower or higher value will hurt the performance.
\vspace{-1.0em}
\paragraph{\textbf{Visualization of Examples}} We further visualize several examples in Fig.~\ref{fig:ssvgexamples} and Fig.~\ref{fig:dvcexamples}. The proposed unified framework achieves improved performance over specialized models (MMN~\cite{wang2021negative} and PDVC~\cite{wang2021end}) on SSVG and DVC tasks, which verifies the effectiveness of the bidirectional prediction tasks.

\end{document}